\documentclass[10pt,twocolumn,letterpaper]{article}

\usepackage{cvpr}
\usepackage{times}
\usepackage{epsfig}
\usepackage{graphicx}
\usepackage{subfigure}
\usepackage{amsmath}
\usepackage{amssymb}
\usepackage{algorithm}
\usepackage{array}
\usepackage{algpseudocode}
\usepackage{bm}
\usepackage{tabularx}

\usepackage[breaklinks=true,bookmarks=false]{hyperref}

\cvprfinalcopy 

\def\cvprPaperID{****} 

\begin{document}

\title{Effective Model Compression via Stage-wise Pruning}

\author{Mingyang Zhang, Xinyi Yu, Jingtao Rong, Linlin Ou\\
College of Information Engineering, Zhejiang University of Technology\\
Hang Zhou, People’s Republic of China\\
{\tt\small linlinou@zjut.edu.cn}
}

\maketitle

\begin{abstract}
Automated Machine Learning(Auto-ML) pruning methods aim at searching a pruning strategy automatically to reduce the computational complexity of deep Convolutional Neural Networks(deep CNNs). However, some previous work found that the results of many Auto-ML pruning methods cannot even surpass the results of the uniformly pruning method. In this paper, the ineffectiveness of Auto-ML pruning which is caused by unfull and unfair training of the supernet is shown. A deep supernet suffers from unfull training because it contains too many candidates. To overcome the unfull training, a stage-wise pruning(SWP) method is proposed, which splits a deep supernet into several stage-wise supernets to reduce the candidate number and utilize inplace distillation to supervise the stage training. Besides, A wide supernet is hit by unfair training since the sampling probability of each channel is unequal. Therefore, the fullnet and the tinynet are sampled in each training iteration to ensure each channel can be overtrained. Remarkably, the proxy performance of the subnets trained with SWP is closer to the actual performance than that of most of the previous Auto-ML pruning work. Experiments show that SWP achieves the state-of-the-art on both CIFAR-10 and ImageNet under the mobile setting.
\end{abstract}
\section{Introduction}

Deep convolutional neural networks(deep CNNs)\cite{b1,b2,b3,b4} have achieved outstanding results in many computer vision tasks. However, deep CNNs comes with a huge computational cost, which limits application on embedded devices(i.e. mobile phone). \\
To expand the application scope of deep CNNs, channel pruning methods were proposed. Traditional channel pruning methods always rely on human-design rules\cite{b6,b7}. Recently, inspired by the Neural Architecture Search(NAS), some AutoML-based pruning work\cite{b8,b9,b12} has been proposed to automatically prune channels without a human-design mode. Considering a network with 10 layers(each layer contains 32 channels), the candidates of each layer and the whole network could be $32$ and $32^{10}$, respectively. Thus, AutoML-based pruning methods can be seen as fine-grained NAS because of more candidates than normal NAS\cite{b28,b27,b23} in each layer. \\
For above mentioned AutoML-based methods, some based on reinforcement learning or evolutionary algorithm\cite{b19,b8,b7} are quite time-consuming due to iterative retrain for each pruned network. To reduce the computation in pruning, some AutoML-based work\cite{b35,b10,b9,b12} shares the weights through training a supernet for all candidate pruned networks which are called subnets. A typical weight-sharing pruning approach contains three steps: training a supernet by iteratively sampling and updating different candidates, searching the best subnet based on the evolutionary or greedy algorithm and training the best subnet from the scratch. However, Chu\cite{b13} considered that the weight-sharing method causes unfull training in the first step since each candidate(subnet) has only a small sampling probability in training. Moreover, unfull training leads to an inaccurate evaluation in the second step, which means some candidates perform well on weight-sharing while bad on training from the scratch. The problem of inaccurate evaluation is particularly obvious in AutoML-based pruning because of more candidates contained in the supernet. Moreover, the width of a supernet also has an impact on the effectiveness of evaluation, which is analyzed in \textbf{Sec. 3.1}\\
To solve the above-mentioned problem, a stage-wise training and searching approach is proposed in this paper. Inspired by \cite{b14}, we consider to divide a deep supernet into several stage-wise supernet(i.e. ResNet50\cite{b3} consists of 4 stages) for reducing the depth of the supernet. Since each stage of the supernet can be trained and searched independently, the candidate number in one stage is an exponential reduction compared with the whole supernet. With a small search space in each stage, the sampling probability of the candidates is raised, which means that each supernet can be fully trained. To alleviate the unfair training result caused by the width of the supernet, both the fullnet and the tinynet, which has the largest and smallest width respectively, are trained in each iteration. Besides, we present a distributed evolutionary algorithm where each stage can be searched independently in terms of an evolutionary algorithm(EA). The constraints(i.e. FLOPs, latency) for each EA are provided by another EA, called EA manager, where the EA manager searches for the best combination of FLOPs in each stage. Due to small and independent stage-wise search space, each EA can be sped up in a parallel way.\\
However, the stage-wise supernet in each stage cannot be trained without internal ground truth. To solve such the problem, an existing pre-trained neural network was used to generate stage-wise feature maps that are viewed as ground truth in each stage. Nevertheless, It is also time-consuming to obtain a pre-trained neural network as the teacher network. Besides, the structural difference between the teacher network and the student networks has a strong impact on distillation results. Hence, a stage-wise inplace distillation method is put forward for the fullnet(the largest width subnet) to supervise the learning of subnets. It is worth noting that the fullnet is jointly trained with other subnets. Thus, there does not exist an extra cost for obtaining the fullnet.\\
Our contribution lies in four folds:
\begin{figure*}[htbp]
\centering
\subfigure[]{
\begin{minipage}{0.4\linewidth}
\includegraphics[width=1.0\linewidth]{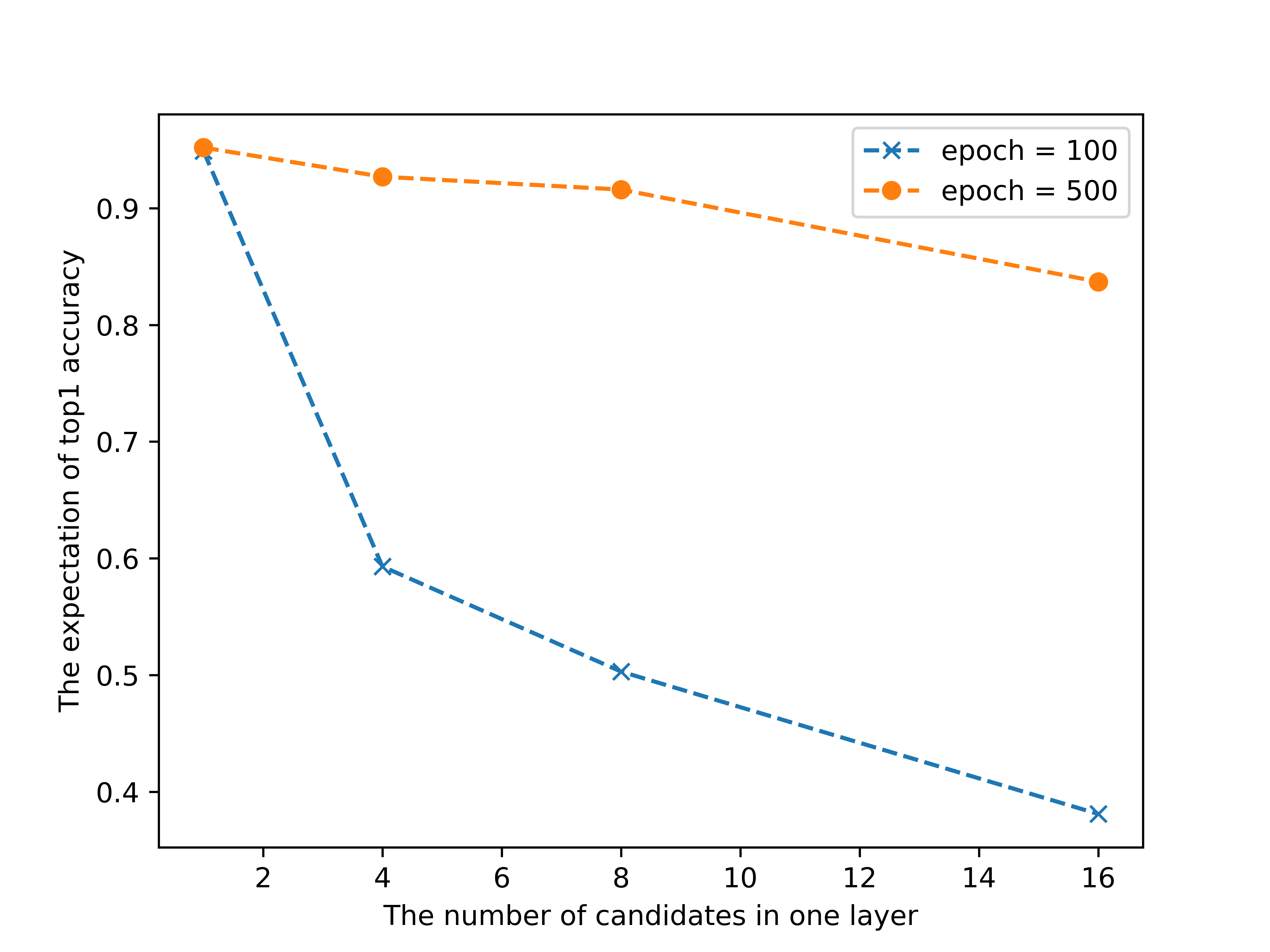}
\end{minipage}}
\subfigure[]{
\begin{minipage}{0.4\linewidth}
\includegraphics[width=1.0\linewidth]{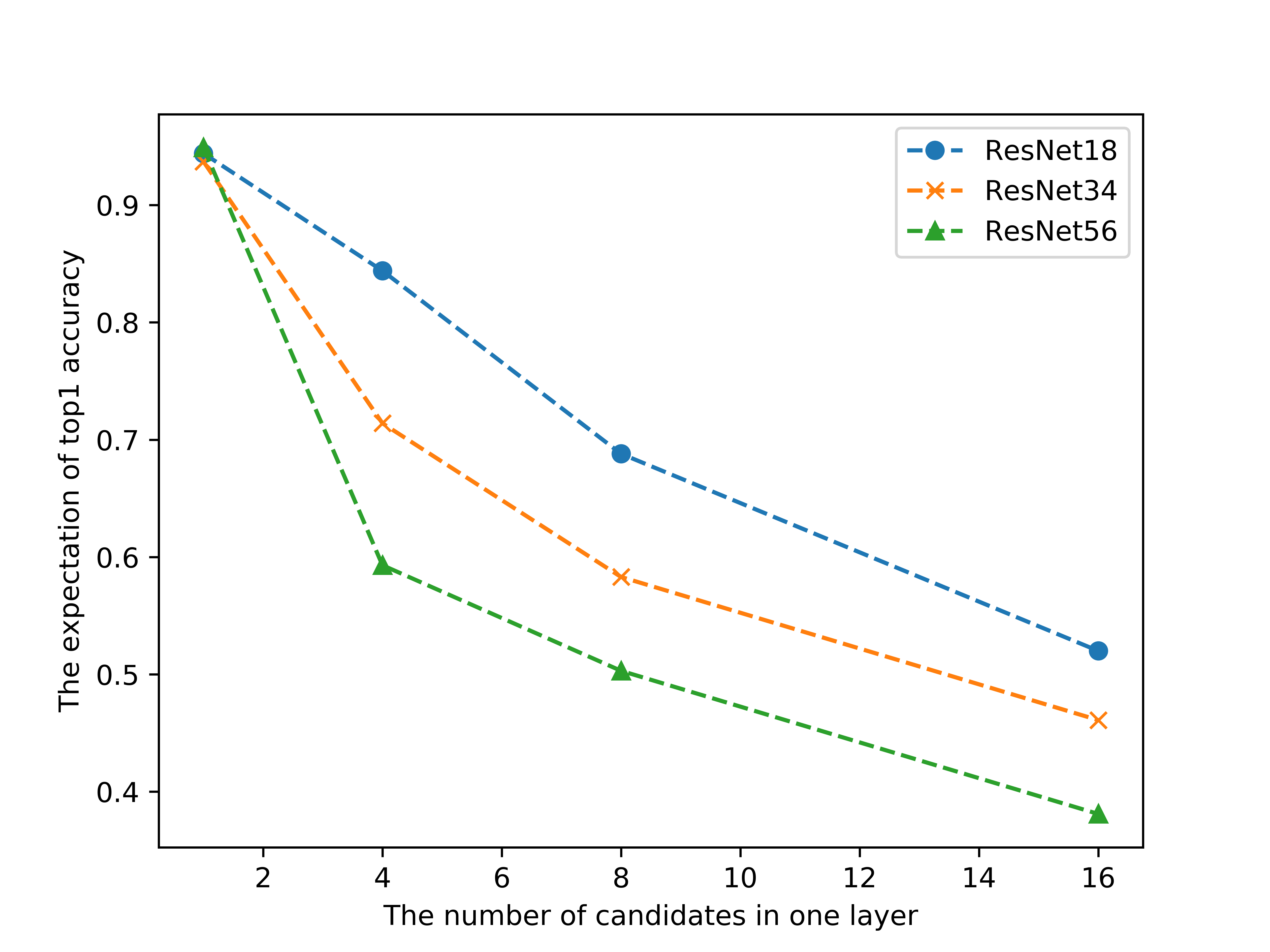}
\end{minipage}}
\caption{(a) The expectation of top-1 accuracy collected from ResNet56\cite{b3} with different candidate number in one layer. The blue and red dash line denotes ResNet56 trained on CIFAR10\cite{b33} under 100 epoches and 500 epoches, respectively. (b) The expectation of top-1 accuracy collected from ResNet18\cite{b3} , ResNet34\cite{b3}  and ResNet56 with different candidate number in one layer. All models are trained under 100 epoches on CIFAR10\cite{b33}.}
\label{1}
\end{figure*}
\begin{itemize}
   \item We propose a stage-wise training and searching pipeline for both channel pruning and NAS. By splitting a CNN into several stages, the number of stage-wise candidates is exponential reduction in contrast to the net-wise candidates. Hence, each candidate obtains full training, which is the essence of accurate evaluation for searching.
  \item To conveniently provide stage-wise ground truth for each stage, a stage-wise inplace distillation method is presented through the joint training of the fullnet and subnets. Thus, the fullnet can easily supervise the learning of the subnets by offering stage-wise feature maps.
  \item To accelerate the searching process, a distributed evolutionary algorithm is suggested. Each stage can be searched by the EA with constraints given by an EA manager in a parallel way.
  \item Experiments shows that the proposed method can enhance the ranking effectiveness of searching and achieve the state-of-the-art in several datasets. 
\end{itemize}
\section{Related Works} 
\label{sec:section_name}
\textbf{Neural Architecture Search.} The purpose of neural network structure search is to automatically find the optimal network structure with reinforcement learning(RL)\cite{b38,b21}, evolutionary algorithm(EA)\cite{b22}, gradient \cite{b23,b24,b25} and parameter sharing methods\cite{b26,b27,b28}. RL-based and EA-based methods need to evaluate each sampled network by retraining them on the dataset, which is time-consuming. The gradient-based method can simultaneously train and search the optimal subnet through assigning a learnable weight to each candidate operation. However, the gradient-based approach causes unfair training results because some candidates obtain more learning resources than others\cite{b13}. Moreover, since gradient-based approaches need more memory for training, it cannot be applied to the large-scale dataset. Parameter sharing methods can search on the large-scale dataset by only activating one candidate in each training iteration. Nevertheless, parameter sharing methods cause unfull training results\cite{b13}. Unfull or unfair training results will cause an inaccurate evaluation of searching, which means that the best-searched architecture is not the optimal one after retraining. To solve such a problem, Li\cite{b14} proposed a blockwisely searching method, which can more fully train each sampled subnets. \\
\textbf{Pruning for CNNs.} Pruning some redundant weights is a prevalent method to accelerate the inference of CNNs. According to the different granularity of pruning, it is divided into weight pruning and channel pruning. For weight pruning, the individual weights in the channel are removed based on some rules\cite{b29}, which causes unstructured sparse filters and cannot be accelerated directly on most hardware. Therefore, much recent work focus on channel pruning. Channel pruning methods \cite{b30,b32,b19,b18,b6} can accelerate the inference of CNNs on general-purpose hardware by reducing the number of filters since the remaining filters are structural. Though the above methods achieve remarkable improvement in the practicality of pruning, it still needs human-designed heuristics to guide pruning.\\
\textbf{Auto-ML pruning.} Recently, inspired by NAS work, AutoML pruning methods\cite{b8,b9,b12,b35,b10} have attracted growing interest in automatically pruning for deep CNNs. Different from NAS, the candidate choices are consecutive in the channel pruning task. Compared with pruning methods based on the human-craft rule, AutoML pruning methods aim to search for the best configuration without manual tuning. AMC\cite{b8} adopted a DDPG agent to sample a pruned network. And the performance of the pruned networks is evaluated by training from the scratch, which is time-consuming and cannot be applied to a large-scale dataset. MetaPruning\cite{b9} trained a PruningNet that can predict weights for any pruned networks, while the parameter amount of the PruningNet is several times of the target network, which leads to unfull train. AutoSlim\cite{b12} first trained a slimmable network\cite{b35} where the weights between different widths are shared through the supernet, and then searched the best subnet in terms of the greedy algorithm. However, the width of the convolutional layer in each subnet must be the same in training. This leads to the problem that the best subnet achieves the highest accuracy with weight sharing, but at the same time, the best subnet gets poor performance when trained from the scratch. To keep the consistency of searching and retraining results, the proposed stage-wise pruning method splits a CNN into several stages and trains them separately under the supervision of the fullnet, which will be explained in Sec.3.\\
\textbf{Knowledge Distillation.} Knowledge distillation is used to train the small student model on a transfer feature set with soft labels or intermediate representations provided by the large teacher model. Soft targets lead to the superior performance of knowledge distillation\cite{b45}. However, as the network is designed to become deeper and deeper, it is not enough to just transfer knowledge to a student network from a teacher network by soft targets. To solve such a problem, some previous work\cite{b46,b47,b48,b49,b50} transferred the knowledge for the student network from the internal representation of the teacher network. All existing work assumed that the teacher network has been pretrained. Nevertheless, it is always time-consuming to train from scratch to obtain a pretrained teacher network that is essential for various knowledge distillation methods. For example, it may cost more than 10 GPU days to train a ResNet on ImageNet. Moreover, Liu\cite{b20} found that the architecture of the teacher and the student networks has a huge impact on transferring results. Hence, we proposed a stage-wise inplace distillation method to overcome the gap and to reduce the time consumption.
\begin{figure*}[t]
\centering
\includegraphics[width=1.0\linewidth]{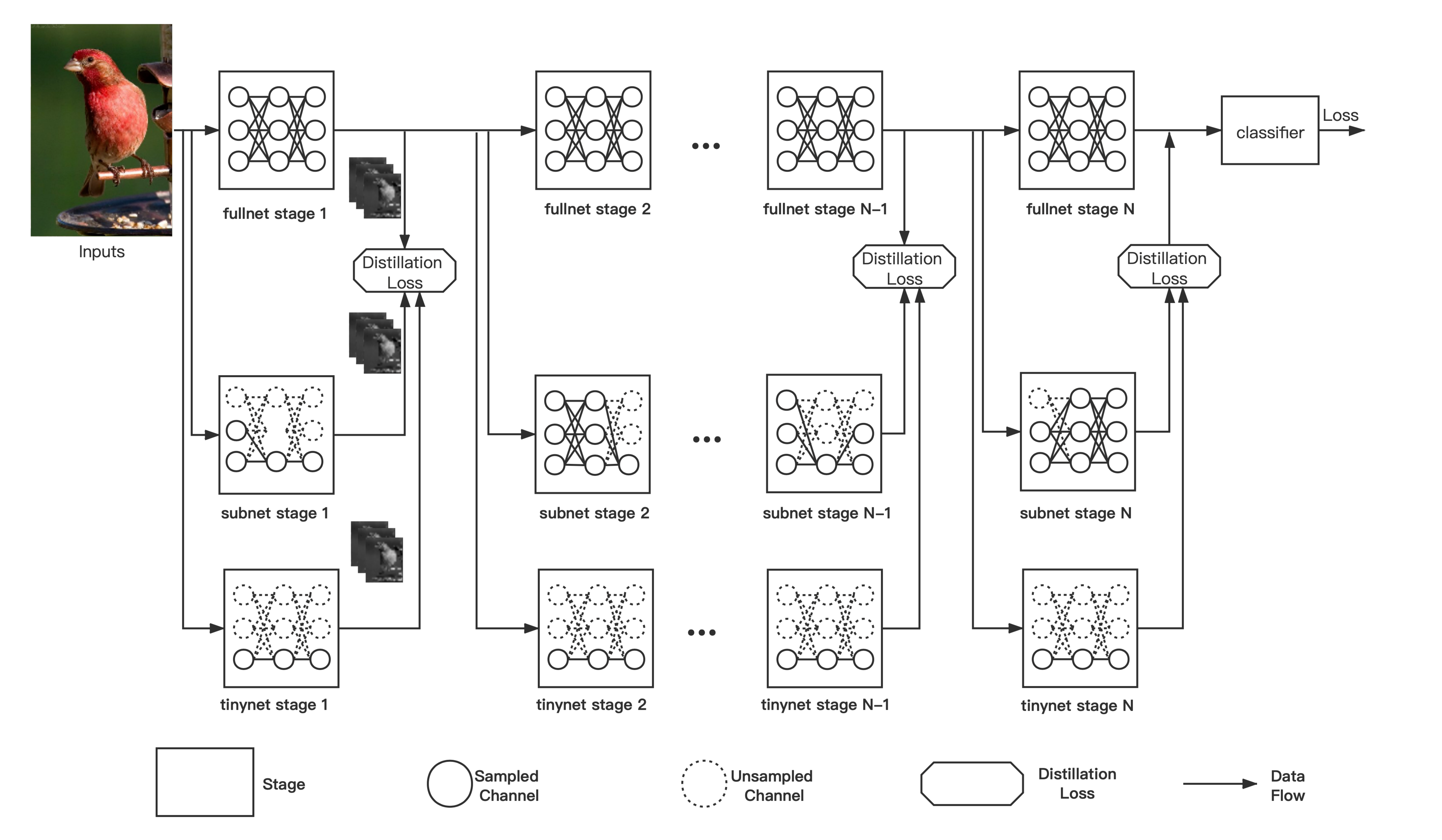}
\caption{Illustration of the stage-wise training. There are three forms of the network, including the fullnet, the subnet and the tinynet. The fullnet infers inputs once to generate and transfer its knowledge to the subnet and the tinynet by minimizing the L2-distance between the their stage-wise output feature maps. It is worthy to note that these three networks are weight sharing.}
\label{2}
\end{figure*}
\section{Stage-wise Pruning}
The problem of inaccurate evaluation caused by weight sharing is introduced in \textbf{Sec. 3.1}. We find that the depth and width of the supernet have an impact on training adequacy. Thus, the stage-wise inplace-distillation is proposed in \textbf{Sec. 3.2} to alleviate the aforementioned drawbacks. To efficiently search the optimal subnet from supernet trained by the stage-wise inplace-distillation, we present a distributed evolutionary algorithm in \textbf{Sec. 3.3.}
\subsection{Challenge of Weight sharing} 
\label{sub:challengeofweightsharing}
AutoML pruning methods always need to train a supernet that shares weights for all subnets firstly and almost immediately evaluates the accuracy for each subnet. For many AutoML pruning approaches \cite{b8,b9,b12}, pruning candidates(subnets) directly compare with each other according to evaluation accuracy. The subnets with higher evaluation accuracy are selected and expected to deliver high accuracy after training from the scratch. However, such an intention cannot be necessarily achieved since some subnets which have outstanding performance on shared parameters perform poorly after training from scratch. \\
To visualize the performance drop of weight-sharing, we train a supernet with different candidate numbers. For a trained supernet, we randomly sample a batch of subnets from the supernet and evaluate them on the validation dataset. Statistical accuracy expectation $E(a_{super})$ is utilized to evaluate whether the supernet is adequately trained. $E(a_{super})$ is written as 
\begin{equation}
  E(a_{super})= \sum _ { i = 1 } ^ { n }  a_{sub_i}
\end{equation}
where $n$ denotes the number of the randomly sampled subnet and $a_{sub_i}$ represents the accuracy of $i$th subnet. As shown in Figure \ref{1}(a), with the increase of the candidate number, the top-1 accuracy expectation of supernet dramatically degrades under 100 epochs while it falls slightly under 500 epochs. Moreover, we train three different depth supernets and then calculate their $E(a_{super})$s. It is found that the expectation of top-1 accuracy is related to the depth of the CNN, which is shown in Figure \ref{1}(b). Next, the problem of weight sharing in terms of depth and width is discussed.\\ 
\textbf{The depth of supernet.} CNNs(i.e. ResNet152\cite{b3}) is designed deep to enhance the representative ability, which exponentially increases the subnet number in pruning. The subnet number $N$ that inherit weights from the supernet can be formulated as 
\begin{equation}
  ||N|| = g^L
\end{equation}
where $g$ denotes the candidate number for each convolutional layer and $L$ represents the depth of the CNN. For the channel pruning of a deep CNN, the search space $N$ is always large(e.g., $>30^{50}$). Hence, many subnets get unfull training results due to weight sharing, which leads to the ineffectiveness of evaluation. \\
\textbf{The width of layers.} Some Auto-ML pruning work\cite{b12,b9,b35,b11,b10} train a single neural network executable at different widths as the supernet. There is not only cross-layer weight sharing but also within the layer. In one layer, the parameters of different widths(candidates) are shared. For instance, all parameters of 0.25$\times$(width scaled by 0.25 of the original width) are shared with the half parameters of 0.5$\times$. Each sampling on any training step is independent of each other. Thus, for $n$ training steps, the sampling times expectation of $i$th channel can be formulated as 
\begin{equation}
  E(c_i) = (1 - \frac{i-1}{m})n
\end{equation}
where $m$ denotes the channel number in the layer. Therefore, the channels with small indexes can be trained more times, which causes unfair training results. Formally, we consider a common supernet that contains $L$ layers, each with $m$ channels. In pruning, a group of ratio sequence $R$ can be obtained under certain constraint(i.e. FLOPs), where each ratio sequence $r = [c_1, ..., c_L] \in R$. Because of independent sampling in each layer, the sampling times expectation of $r$ can be described as  
\begin{equation}
  E(r) = E(c_1)E(c_2)...E(c_L)
\end{equation}
According to the inequality of arithmetic and geometric means, we have
\begin{equation}
  E(r) \le (\frac{\sum_{i=1}^{L}E(c_i)}{L})^L
\end{equation}
Equality holds if and only if $c_1=c_2=...=c_L=c$. Hence,
\begin{equation}
  E(r) \le ((1-\frac{cm-1}{m})n)^L
\end{equation}
That is to say, the subnet with uniform sampling obtains the most training resource under certain constraint, which means that the pruning strategy of previous work always prone to get a uniform sampling model.\\ 
\subsection{Stage-wise Inplace Distillation for Training} 
\label{sub:subsection_name}
\begin{figure*}[t]
\centering
\includegraphics[width=0.8\linewidth]{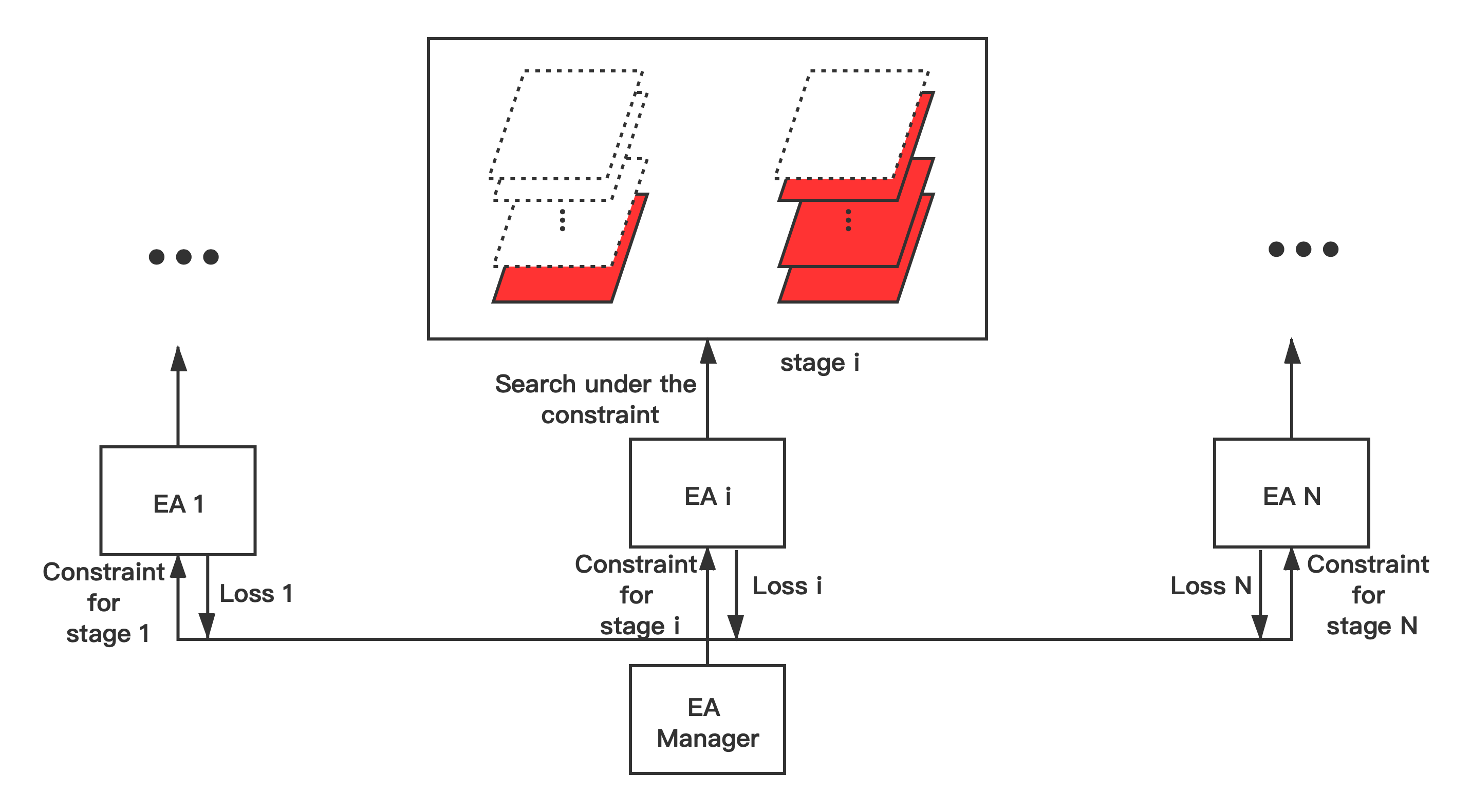}
\caption{Illustration of the distributed evolutionary algorithm. Both EA Manager and EA are modified from an evolutionary algorithm. Given FLOPs constraint for the whole network, EA Manager is responsible for searching for the best combination of stage-wise FLOPs. The feedback of each FLOPs gene in EA Manager is provided by each EA with searching for the smallest distillation loss under the stage-wise FLOPs constraint.}
\label{3}
\end{figure*}

As mentioned before, too many candidates in training can lead to ineffective evaluation on searching because of unfull training results. To adequately train the supernet, we divide the supernet $S$ into N stages according to the depth. Hence, the search space of supernet $S$ can be represented as
\begin{equation}
  S = [S_1 ,..., S_{i}, S_{i+1}, ..., S_N]
\end{equation}
where $S_i$ denotes the stage-wise supernet of $i$th stage. Then we can train the supernet by training the stages separately. The learning of the stage $i$ can be formulated as
\begin{equation}
  W_i^* = min_{W_i} L_{train}(W_i, S_{}i; X, Y)
\end{equation}
where X and Y denote the input data and the groud truth labels, respectively. Subsequently, the candidates number for $i$ th stage can be written as
\begin{equation}
  ||S_i|| = g^{L_i}
\end{equation}
where $L_i$ denotes the depth of the $i$th stage and it is smaller than $L$. The search space can be extremely reduced when we train stage-wise supernet in each stage independently. \\
However, internal ground truth in Eq.(8) cannot be obtained directly from the dataset. One solution is using block-wise feature maps generated by a pre-trained network to supervise the training of subnets. However, it is time-consuming to obtain a pre-trained network through training from the scratch in practice(e.g. ResNet50 $>$ 10 GPU days). Besides, the architecture of teacher and student networks has a huge impact on transferring results\cite{b20}. \\
To tackle the above problem, the stage-wise inplace distillation is proposed here. The essential idea behind the inplace distillation\cite{b10} is to transfer knowledge inside a single supernet from the fullnet to a subnet in each training iteration. For an individual convolutional layer, the performance of the wider candidate cannot be worse than the slim one. Because the wider one can achieve the performance of the slim one by learning weights from some unuseful channels to zeros. Therefore, the performance of any candidates is bounded as follows
\begin{equation}
 |y^{f}-y^{f}|\le |y^{f}-y^{r}| \le |y^{f}- y^{s}|
\end{equation}
where $y^{r}=\sum_{i=1}^{r} w_ix_i$ is the aggregated feature, $r$, $s$ and $f$ denote the channel number of random sampled, the smallest and the largest candidates, respectively. The rule in Eq.(10) can also be extended to the whole supernet, which means the performance of the subnet with any width is bounded in the tinynet and the fullnet.\\ 
In stage-wise inplace distillation, we use the stage-wise representation of the fullnet to supervise the training of subnets. The pipeline of the supervision with stage-wise inplace distillation is shown in Figure \ref{2}. The output $\hat{Y}_{i-1}$ of the $(i-1)$th stage from the fullnet is adopted by the input of the $i$th stage of subnets. To supervise the subnets learning from the fullnet, the following MSE loss is considered as the distillation loss in Figure \ref{3}
\begin{equation}
  L_{train}(\hat{Y}_{i-1}, Y_i) = \frac{1}{F}||Y_i-\hat{Y_i}||_2^2
\end{equation}
where $Y_i$ and $\hat{Y_i}$ denote the output of the subnets and fullnet in $i$th stage, respectively, $F$ is the number of the channels in $Y$. \\
As mentioned in \ref{sub:challengeofweightsharing}, the channels with larger indexes suffer from unfull training. To ensure the sufficient training of each channel, an intuitive approach is overtraining. Given a batch of input images and ground truth labels, we first calculate the task loss(e.g. cross-entropy) and the gradients of fullnet through forward and backward propagation, simultaneously. The stage-wise feature maps of fullnet $\hat{Y} = [\hat{Y}_1, ... \hat{Y}_N]$ are saved. Subsequently, under the supervision of $\hat{Y}$, the distillation loss in Eq.(11) and the gradients of the stage-wise subnet are calculated. Furthermore, in a subnet training process, we train the smallest width(tinynet) to improve the performance of the supernet. Based on this, each channel can be trained at least once in one iteration. The ground truth label has been generated in the fullnet training process. Thus, the training of each stage-wise subnet can be sped up in a parallel way. The detailed algorithm is described in Algorithm \ref{alg:train algorithm}. We use \textbf{multiprocessing} module\footnotemark[1] to parallelize our algorithm.
\footnotetext[1]{The multiprocessing module is applied to start the child process and execute our customized tasks in the child process. https://github.com/python/cpython/tree/3.9/Lib/multiprocessing/}
\begin{algorithm}[ht] 
\caption{ Framework of supervision with stage-wise inplace distillation.} 
\label{alg:train algorithm} 
\begin{algorithmic}[1] 
\Require 
The fullnet $S$, the stage-wise supernets $[S_1,...,S_N]$ and the dataset $(X,Y)$;
\Ensure 
The well-trained stage-wise supernets $[S_1,...,S_N]$; 
\State \textbf{for} $t=1,...,T$ \textbf{do}
\State \hspace*{1em} Get next mini-batch of data $x$ and label $y$ from $(X,Y)$  
\State \hspace*{1em} Execute fullnet $y'=S(x)$, and save stage-wise feature maps $\hat{X}=[x,...,\hat{Y}_{N-1}]$, $\hat{Y}=[\hat{Y}_1,...,\hat{Y}_N]$ 
\State \hspace*{1em} Caculate cross entropy loss, $loss = CE(y', y)$
\State \hspace*{1em} Clear gradients, $optimizer.zero\_grad()$
\State \hspace*{1em} Accumulate gradients, $loss.backward()$
\State \hspace*{1em} Randomly sample width for convolutional layers and obtain stage-wise subnets, $S_r=[S_{r1},...,S_{rN}]$
\State \hspace*{1em} Uniformly sample smallest width for convolutional layers and obtain stage-wise tinynets, $S_t=[S_{t1},...,S_{tN}]$
\State \hspace*{1em} \textbf{multiprocessing} $s=[S_r, S_t]$, $x_s=[\hat{X},\hat{X}]$, $y_s=[\hat{Y}, \hat{Y}]$ \textbf{do}
\State \hspace*{1em} \hspace*{1em} Execute subnet, $y'=s(x_s)$
\State \hspace*{1em} \hspace*{1em} Caculate distillation loss, $loss = L(y', y_s)$
\State \hspace*{1em} \hspace*{1em} Accumulate gradients, $loss.backward()$
\State \hspace*{1em} \textbf{end multiprocessing}
\State \hspace*{1em} Update weights, $optimizer.step()$
\State \textbf{end for}\\
\Return trained stage-wise suernets $[S_1,...,S_N]$; 
\end{algorithmic} 
\end{algorithm}
{}
\subsection{Distributed Evolutionary for Searching} 
\label{sub:distributed_evolutionary_for_searching}
After the stage-wise supernet is trained, the learning ability of a subnet can be evaluated by its loss in each stage. However, each stage-wise supernet still contains about  $30^{10}$ stage-wise subnets. It is infeasible to evaluate all of them. For previous one-shot pruning methods, randomly sampling, EA-based and RL-based methods have been used to sample sub-models from the trained supernet for further evaluation. The most recent work found that EA can search better model than RL but need to spend more time on searching. For the proposed stage-wise inplace distillation, a novel method is suggested to search the best subnet according to the stage-wise performance under certain constraint.\\
Because of inplace distillation training mentioned above, the EA in Figure \ref{3} is applied to search the best stage-wise subnet that has the smallest distillation loss under given FLOPs constraint. In \cite{b9,b21}, the genes of each stage-wise subnet were encoded with a vector of channel numbers in each layer. Different from the above work, we aim to imitate the behavior of the teacher in each state. Thus, Eq.(11) is used to evaluate the learning ability of each gene. Because the supernet is split into several stages, the search space of an individual EA is shrunk about $10^{60}$ times. Then the top k genes with the lowest loss are selected for generating the new genes with mutation and crossover. The mutation is carried out by changing a proportion of elements in the gene randomly. The crossover means that we randomly recombine the genes in two-parent genes to generate offspring. We can easily enforce the constraint by eliminating the unqualified genes. Through further repeating the top k selection process and new genes generation process for several iterations, the gene that meets constraints while achieving the lowest loss can be obtained. How can the optimal stage-wise constraints be assigned for each stage?\\
To automatically find the best assignment plan of stage-wise constraint, a distributed evolutionary algorithm(DEA) is proposed in this section. The workflow of DEA is revealed in Figure \ref{3}. The EA Manager is also a kind of evolutionary algorithm that provides the strategy of FLOPs constraint for other EAs. Different from EA above, the genes in EA Manager are encoded with a vector of FLOPs constraint in each stage. The evaluation of each gene is the sum of distillation losses given by all stage-wise EAs. Subsequently, the top k genes which are kept generate off-springs in way of mutation and crossover. After several repeating, the optimal stage-wise constraints can be obtained from the top 1 genes. The detail is shown in Algorithm \ref{alg:distributed evolutionary algorithm}. And each stage-wise EA is paralleled to accelerate the searching process. Specifically, we first use the teacher network to generate a batch of representation features for each stage. Therefore, each EA can search for the best stage-wise subnet independently. After searching all of the stages, we can assemble the best model by selecting the best stage-wise subnet from each stage.  
\begin{algorithm}
\caption{ Framework of distributed evolutionary algorithm.} 
\label{alg:distributed evolutionary algorithm} 
\begin{algorithmic}[1] 
\Require 
The constraint $C$, the fullnet $S$, The stage-wise supernets $[S_1,...,S_N]$ and the dataset $(X,Y)$; 
\Ensure 
The best subnet:$S_{top}$;
\State Execute fullnet and save stage-wise feature maps $\hat{Y}$, $y'=S(X)$, $\hat{Y}=[\hat{Y}_1,...,\hat{Y}_N]$
\State Randomly generate a batch of genes $G$ under constraint $C$, $G=[G_1,...,G_N], s.t. ||G_i||=||C_{i1},...,C_{iN}||=C$
\State \textbf{for} $t=1,...,T$ \textbf{do}
\State \hspace*{1em} \textbf{for} $g=G_1,...,G_N$ \textbf{do}
\State \hspace*{1em} \hspace*{1em} Obtain stage-wise constraint from $g$, $g=[C_1,...,C_N]$
\State \hspace*{1em} \hspace*{1em} \textbf{multiprocessing} $c=C_1,...,C_N$, $x_s=X,...,\hat{Y}_{N-1}$, $y_s=\hat{Y}_1,...,\hat{Y}_N$, $s=S_1,...,S_N$ \textbf{do} 
\State \hspace*{1em} \hspace*{1em} \hspace*{1em} Search the best stage-wise subnet $s'$ and caculate distillation loss by $EA$ in Algorithm \ref{alg:evolutionary algorithm}, $(s', L_{p_i})=EA(s,x_s,y_s,c)$ 
\State \hspace*{1em} \hspace*{1em} \textbf{end multiprocessing}
\State \hspace*{1em} \hspace*{1em}Caculate total loss $L$ for $g$, $L=L_{s_1}+...+L_{s_N}$
\State \hspace*{1em} \textbf{end for}
\State \hspace*{1em} Keep top k genes $G_{topk}$ according to $L$
\State \hspace*{1em} Generate $M$ mutation genes, $G_{mutation}=[G_{m1},...,G_{mM}]$, $s.t. ||G_{mi}||=C$
\State \hspace*{1em} Generate $H$ crossover genes, $G_{crossover}=[G_{c1},...,G_{cH}]$, $s.t. ||G_{ci}||=C$
\State \hspace*{1em} Generate new population $G$, $G=G_{mutation}+G_{crossover}$
\State \textbf{end for}
\State Select $S_{top}=[s_1',...,s_k']$ with smallest $L$\\
\Return $S_{top}$; 
\end{algorithmic} 
\end{algorithm}

\begin{algorithm}
\caption{ Framework of evolutionary algorithm.} 
\label{alg:evolutionary algorithm} 
\begin{algorithmic}[1] 
\Require 
The constraint, $C$, the stage-wise supernet $S$, the stage-wise feature maps $(X, Y)$
\Ensure 
The best stage-wise subnet:$S_{top}$ and the stage-wise distillation Loss: $L$;
\State Randomly generate a batch of genes $G$ under constraint $C$, $G=[G_1,...,G_d]$
\State \textbf{for} $t=1,...,T$ \textbf{do}
\State \hspace*{1em} \textbf{for} $g=G_1,...,G_d$ \textbf{do}
\State \hspace*{1em} \hspace*{1em} Construct a stage-wise subnet according to $S$ and $g$, $S_g$
\State \hspace*{1em} \hspace*{1em} Calculate the distillation loss of $S_G$, $L_g = L(S_G(X), Y)$, $L$ from Eq.(11)
\State \hspace*{1em} \textbf{end for}
\State \hspace*{1em} Keep top k genes $G_{topk}$ according to $L_g$
\State \hspace*{1em} Generate $M$ mutation genes under constraint $C$, $G_{mutation}=[G_{m1},...,G_{mM}]$
\State \hspace*{1em} Generate $H$ crossover genes under constraint $C$, $G_{crossover}=[G_{c1},...,G_{cH}]$
\State \hspace*{1em} Generate new population $G$, $G=G_{mutation}+G_{crossover}$
\State \textbf{end for}
\State Select $G_{top}$ with smallest $L_g$\\
\Return $G_{top}$, $L_{G_{top}}$; 
\end{algorithmic} 
\end{algorithm}
\section{Experiments} 
\label{sub:experiments}
In this section, the effectiveness of our proposed stage-wise pruning method is demonstrated. First, We explain the experiment settings on CIFAR-10\cite{b33} and ImageNet 2012 dataset\cite{b15}. Then, we prune ResNet\cite{b3} on CIFAR-10 and visualize the consistency of performance between searching and retraining. Moreover, we apply the stage-wise pruning method to ImageNet 2012 and compare the results with those of other state-of-the-art work. Finally, ablation studies are carried out to find out the influence of using inplace distillation.
\subsection{Setups} 
\label{sub:setups}
The stage-wise pruning method consists of three steps:\\
\textbf{Stage-wise training.} According to resolution size of feature maps, we split ResNet\cite{b3} and MobileNet series\cite{b1,b2} into 4 and 5 stages, respectively. The distillation loss of each stage can be calculated by Eq.(11) . To match the channel number of the fullnet, the output of each stage is connected with a $1\times 1$ convolutional layer without BatchNorm and non-linear activation. As MetaPruning\cite{b9}, the width of each convolutional layer is subdivided into 31 ratios from 0.1 to 1.0.\\
On CIFAR-10\cite{b33} dataset, we randomly sample 200 images for each class from training images as validation dataset. The remaining images are used to train the supernet. We use momentum SGD to optimize the weights, with initial learning rate $\eta = 0.025$, momentum $0.9$, and weight decay $3 \times 10^{-4}$. The supernet is trained for 50 epochs with batch size 512 and the learning rate decays $0.1 \times$ per 10 epochs.\\
On ImageNet 2012\cite{b15} dataset, we randomly sample 50 images for each class from training images as validation dataset. The remaining images are used to train the supernet. We use momentum SGD to optimize the weights, with initial learning rate $\eta = 0.1$, momentum $0.9$, and weight decay $3 \times 10^{-4}$. The supernet is trained for 100 epochs with batch size 512 and the learning rate decays $0.1 \times$ when the epoch is 30, 60 or 90.\\
\textbf{Stage-wise searching.} After training the stage-wise supernet as above, the best subnet is searched in each stage. Firstly, we use the fullnet to generate and save the stage-wise feature maps with 2048 batch size. Subsequently, the hyperparameter of each EA and EA Manager is set to 128 population number, 0.1 mutation probability, 10 iterations. We use 4 and 5 multiprocess to speed up the searching for ResNet and MobileNet series, respectively. Each process can use 2 GPUs to infer with 2048 batch size. \\
\textbf{Retraining} After searching the best subnet, we adopt the same training scheme as \cite{b9} on ImageNet 2012 for both ResNet and MobileNet series. The same lines of \cite{b23} is followed for the training scheme of ResNet on CIFAR-10. It is noted that all baseline models are trained under the same scheme mentioned above.
\subsection{Pruning ResNet on CIFAR-10 and Analysis} 
\label{sub:pruning_on_cifar_10}
To demonstrate the effectiveness of stage-wise pruning, we prune ResNet-56\cite{b3} under 50\% FLOPs constraint on the small dataset of CIFAR-10. As shown in Table \ref{cifar10_results}, our stage-wise pruning method surpasses the baseline model by about 1.4\%. Moreover, our method outperforms all other pruning methods in terms of top-1 accuracy.\\
\begin{table}
\caption{Pruning results of ResNet-56.} 
\label{cifar10_results}
\begin{center}
\begin{tabular}[c]{c|c|c}
\hline
Method&FLOPs(M)&Top-1 Acc(\%)\\
\hline
ResNet-56&125.49&93.27\\
FP\cite{b51}&90.90&93.06\\
RFP\cite{b52}&90.70&93.12\\
HRank\cite{b53}&88.72&93.52\\
EagleEye\cite{b54}&62.23&94.66\\
\textbf{SP(Ours)}&61.36&\textbf{95.03}\\
\hline
\end{tabular}
\end{center}
\end{table}
\begin{figure}
\centering
\includegraphics[width=0.9\linewidth]{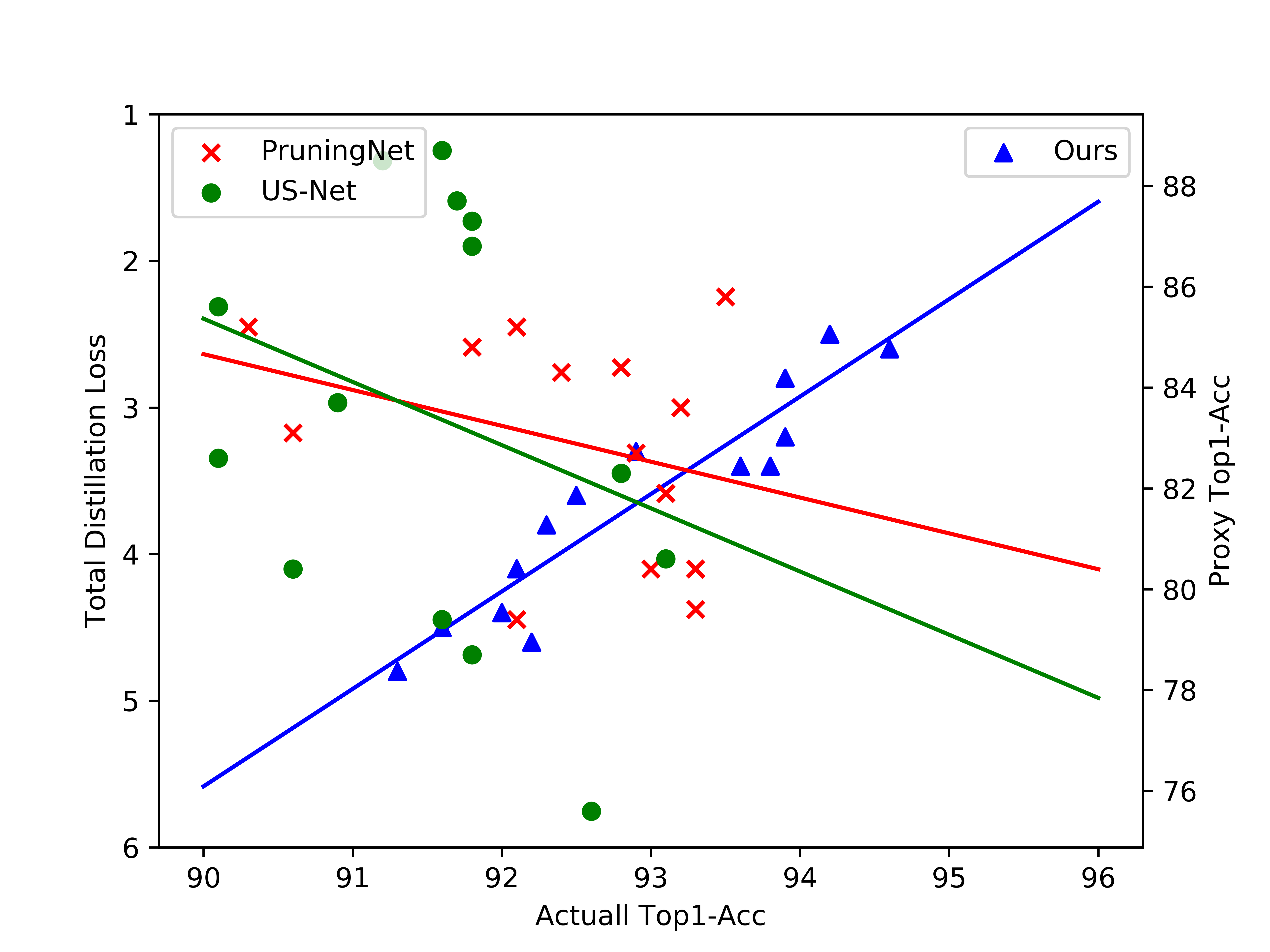}
\caption{Comparison of ranking effectiveness for Stage-wise Pruning(ours), MetaPruning\cite{b9} and AutoSlim\cite{b12}.}
\label{4}
\end{figure}
To evaluate the consistency of model ranking abilities for our method and other AutoML methods, we visualize the relationship between the proxy performance and actual performance. A PruningNet\cite{b9} and a Universally Slimming Network(USNet)\cite{b10} are trained as supernets under the same training scheme due to fairly compare with MetaPruning \cite{b9} and AutoSlim \cite{b12}. The total distillation loss is viewed as the proxy performance of our method. The other two methods take the top-1 accuracy of each subnet that inherits weights from supernet as proxy performance. Each subnet will be trained from scratch in order to obtain its actual performance. As shown in Figure \ref{4}, the proposed method has a strong correlation between the proxy performance and the actual performance while others barely rank the subnets. 
\begin{figure*}[htbp]
\subfigure[]{
\begin{minipage}[t]{0.3\linewidth}
\centering
\includegraphics[width=1.1\linewidth]{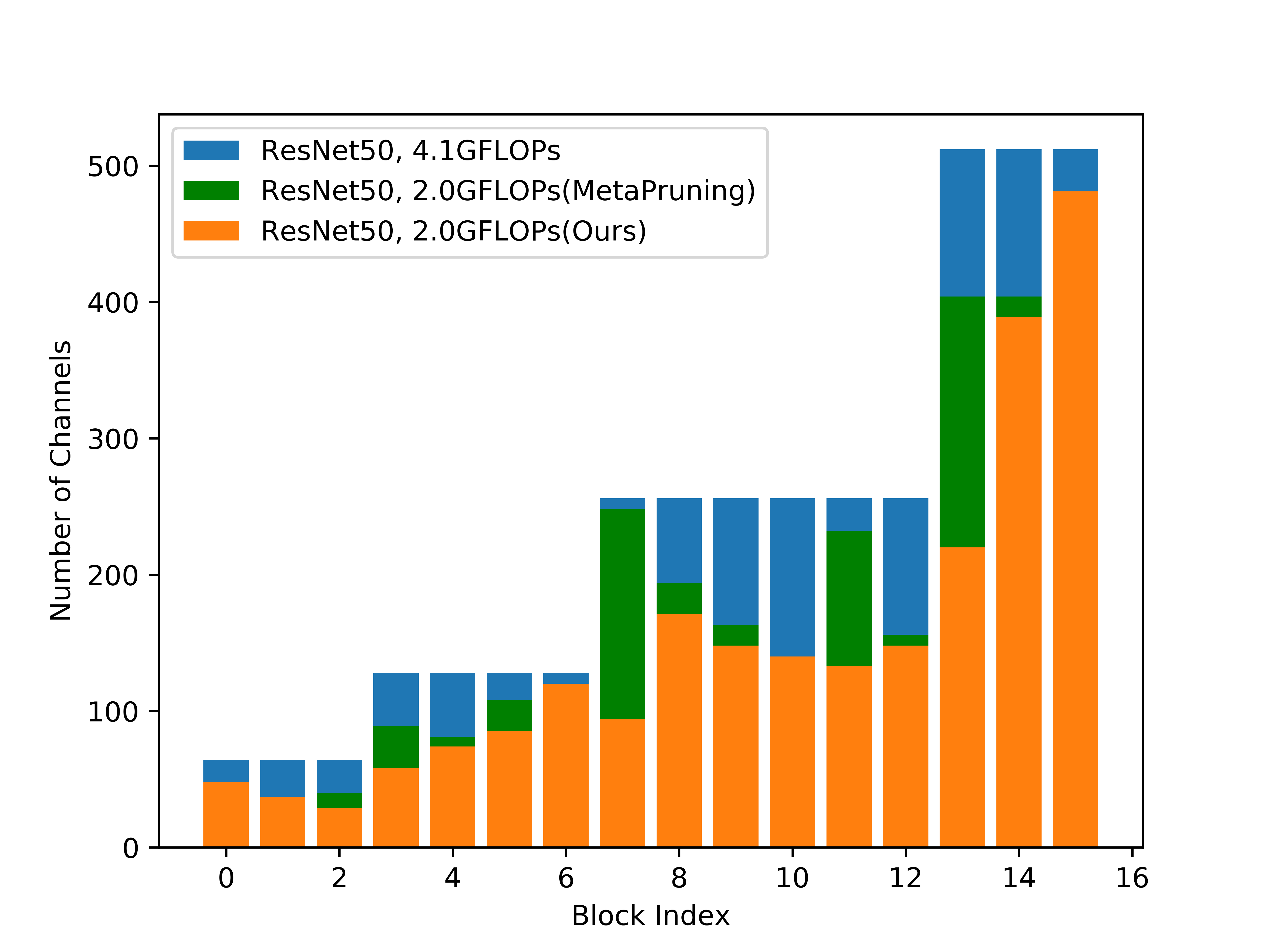}
\end{minipage}}
\subfigure[]{
\begin{minipage}[t]{0.3\linewidth}
\centering
\includegraphics[width=1.1\linewidth]{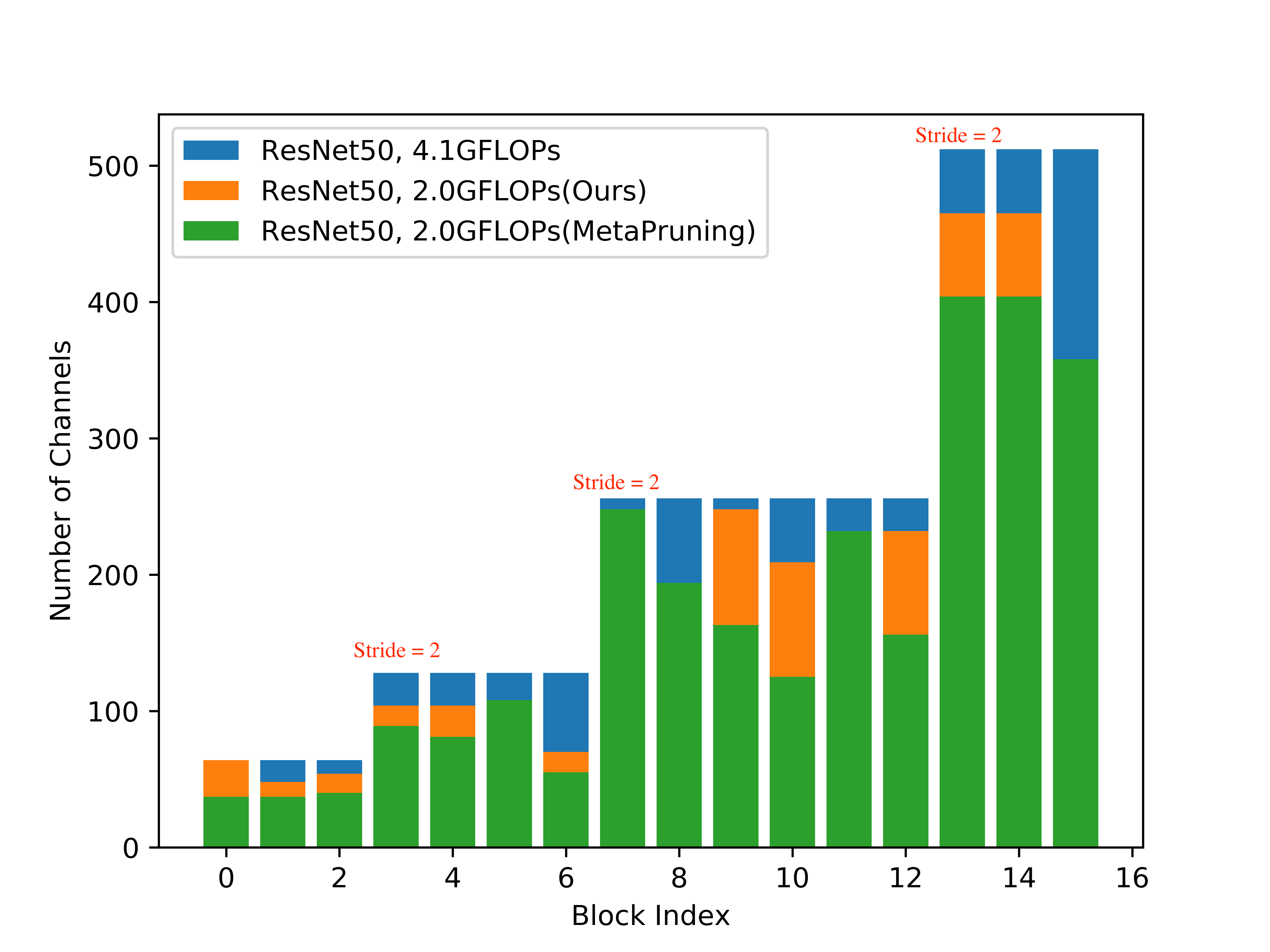}
\end{minipage}}
\subfigure[]{
\begin{minipage}[t]{0.3\linewidth}
\centering
\includegraphics[width=1.1\linewidth]{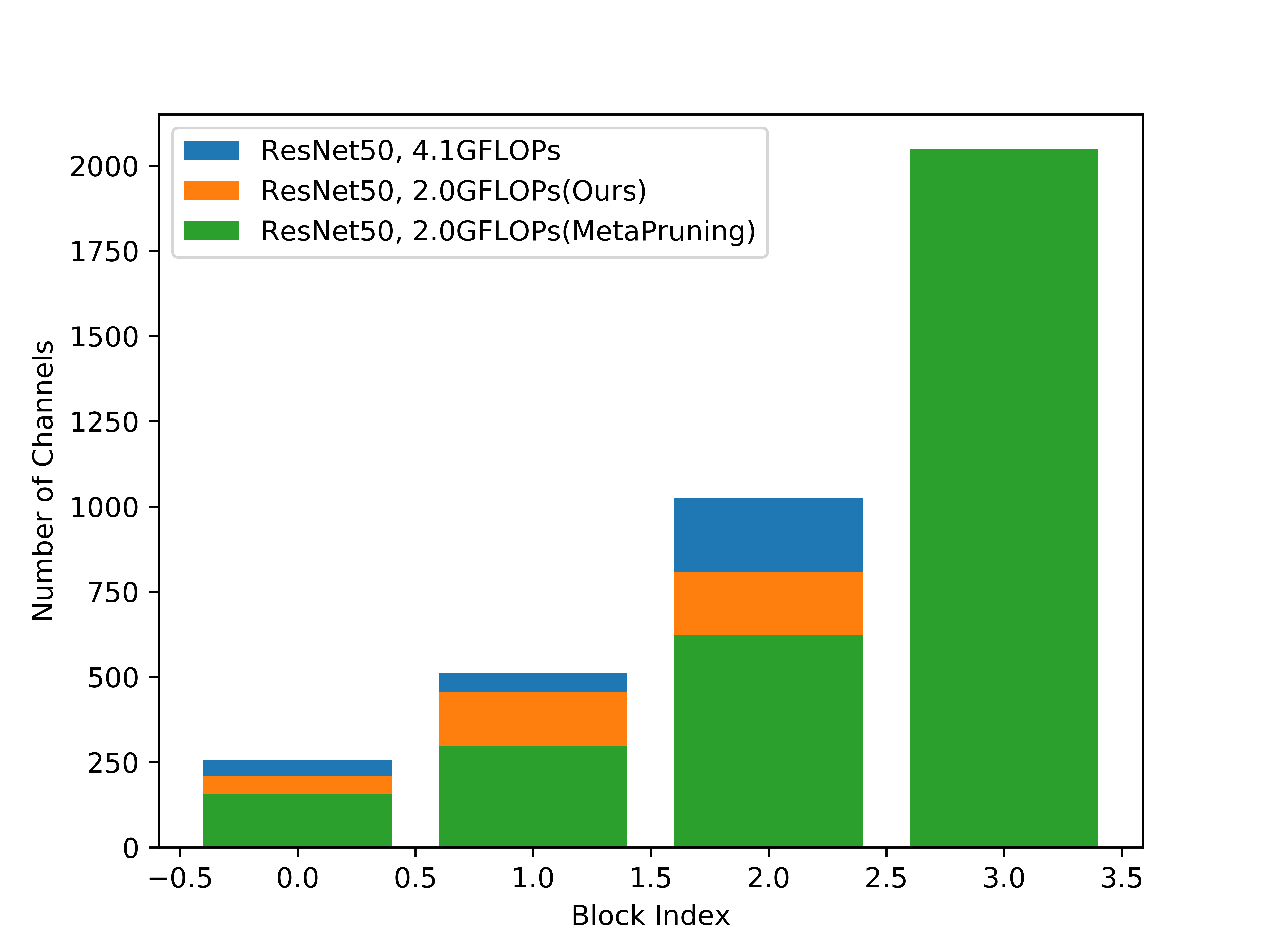}
\end{minipage}}
\caption{The pruning results of ResNet-50. ResNet-50 is stacked by many blocks which consist of three convolutional layers in the main branch. According to the location, we simply divide the three convolutional layers in each block into top layers, middle layers and bottom layers. (a) The number of channels in top layers. (b) The number of channels in the middle layers. (c) The number of channels in the bottom layers.}
\label{5}
\end{figure*}
\subsection{Pruning MobileNet and ResNet on ImageNet 2012} 
\label{sub:pruning_mobilenet_and_resnet}
\begin{figure*}[htbp]
\subfigure[]{
\begin{minipage}[t]{0.3\linewidth}
\centering
\includegraphics[width=1.1\linewidth]{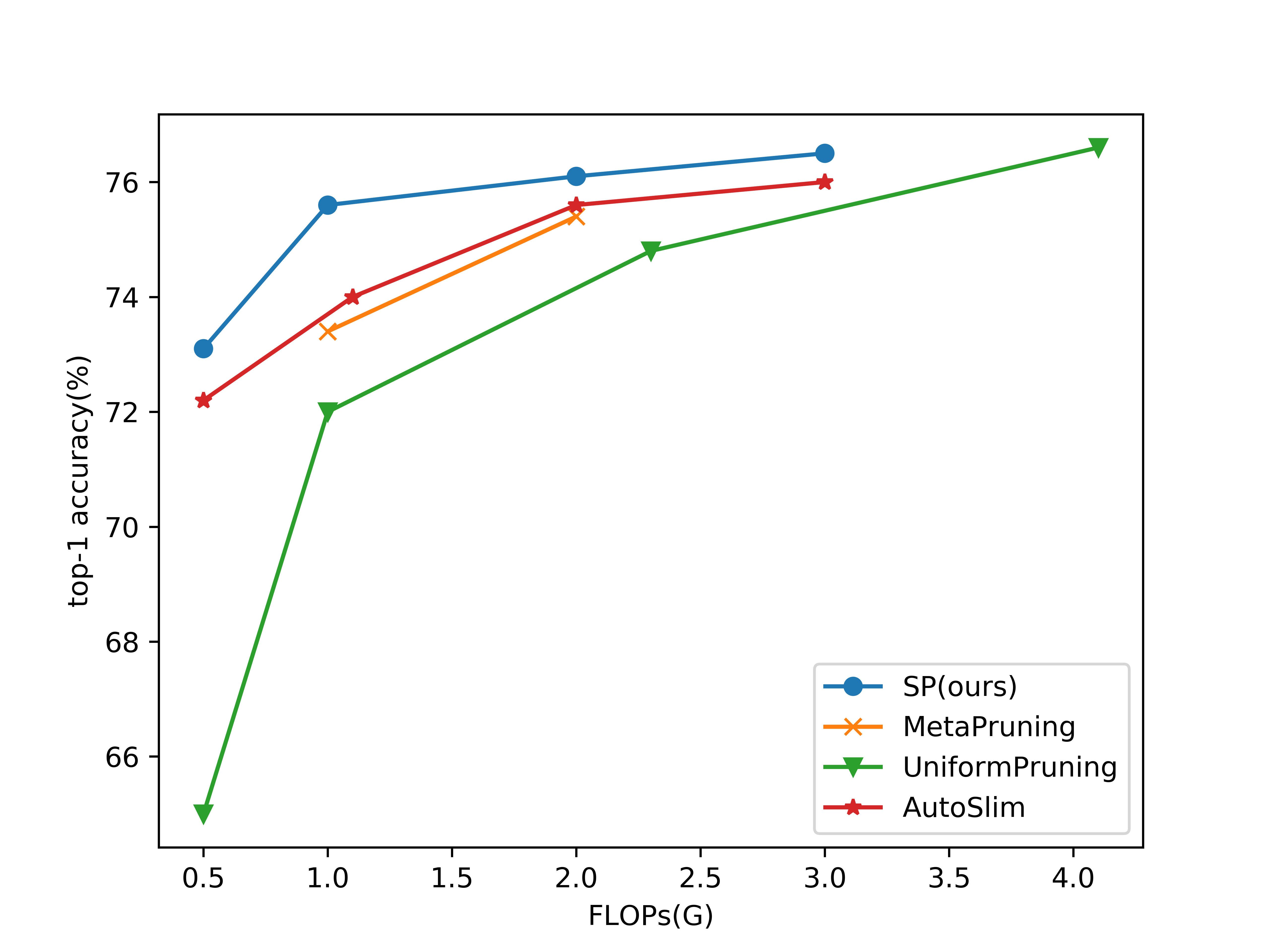}
\end{minipage}}
\subfigure[]{
\begin{minipage}[t]{0.3\linewidth}
\centering
\includegraphics[width=1.1\linewidth]{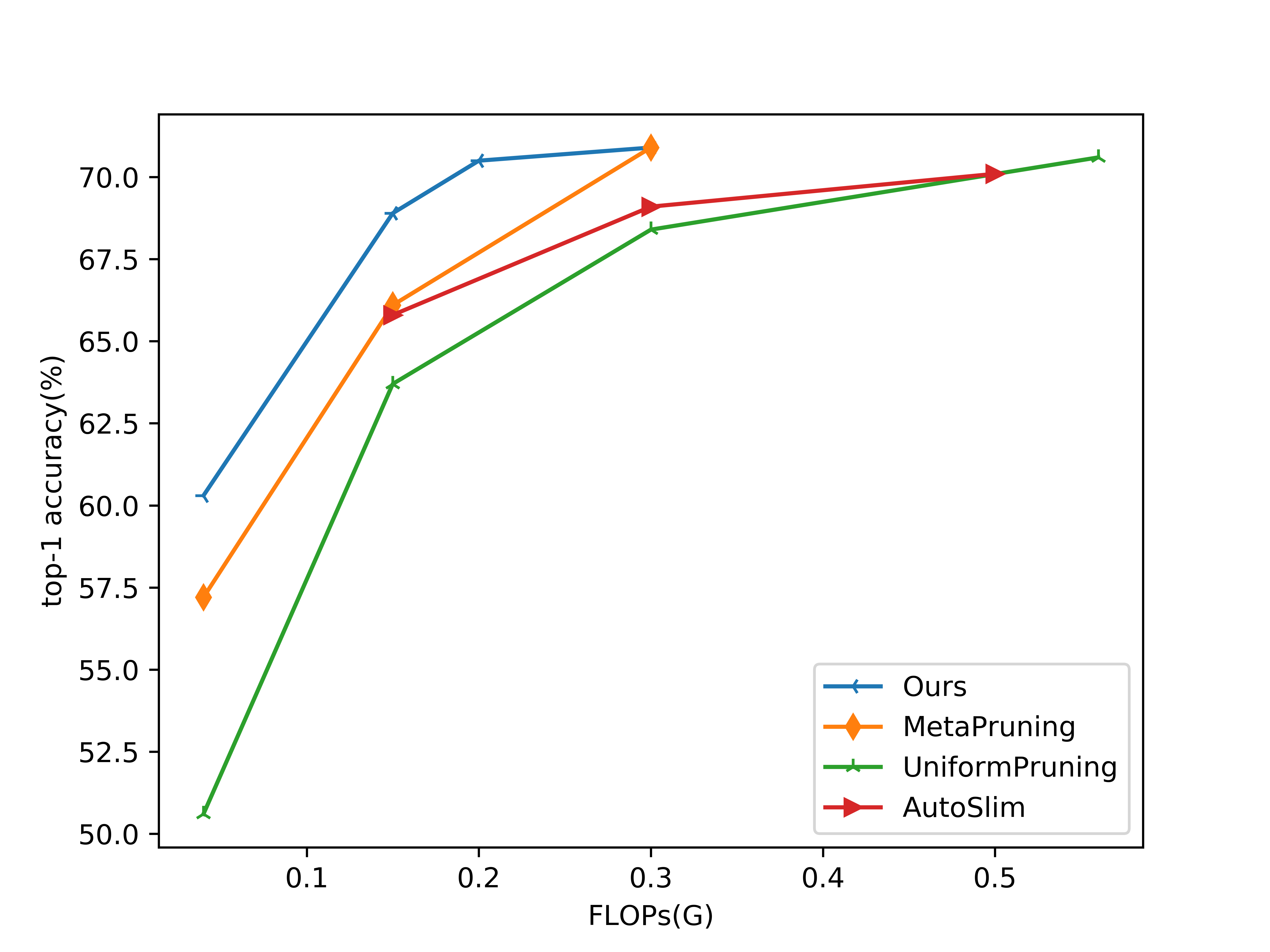}
\end{minipage}}
\subfigure[]{
\begin{minipage}[t]{0.3\linewidth}
\centering
\includegraphics[width=1.1\linewidth]{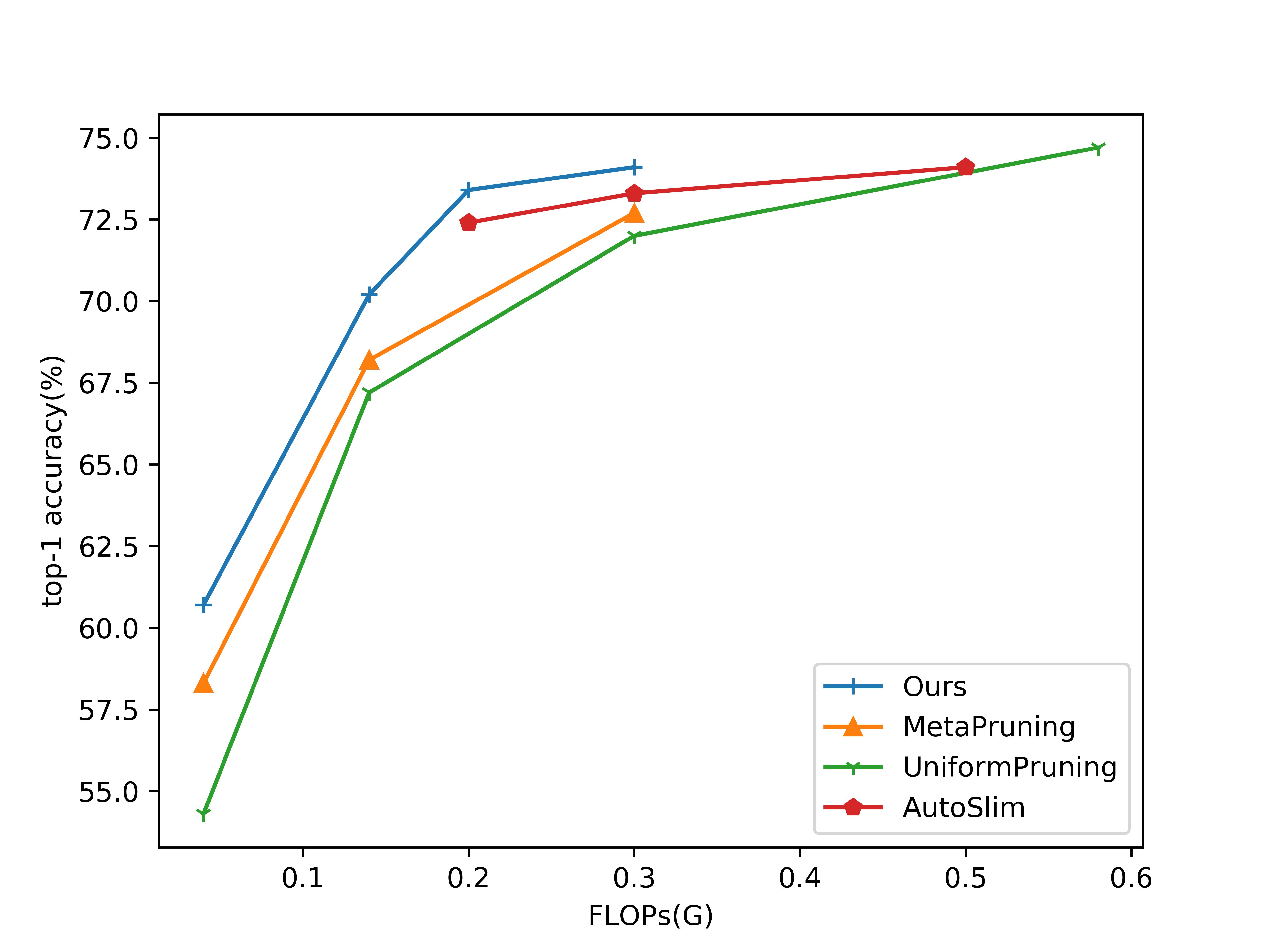}
\end{minipage}}
\caption{The accuracy-FLOPs tradeoff of Auto-ML pruning methods on (a) ResNet-50, (b) MobileNetV2, (c) MobileNetV1. All models are pruned on ImageNet 2012.}
\label{6}
\end{figure*}
In addition, our method is extended to lighting models on a large-scale dataset, ImageNet 2012. Table \ref{imagenet_results} summarizes our results on MobileNet V1\cite{b1}, MobileNet V2\cite{b2} and ResNet-50\cite{b3}. It is noted that we experiment with both residual and nonresidual networks. We compare our results with uniformly pruned baselines and other recent channel pruning methods. It is shown that our method achieves the best results across different computational budgets. In the case of extreme pruning(i.e. 40M FLOPs), MobileNetV1/MobileNetV2 pruned by SWP outperform baseline model to a considerable degree(9.7\% and 6.4\%). Figure \ref{5} compares the curve of top-1 accuracy and FLOPs for the most recent Auto-ML pruning methods and uniformly pruning method. Our SWP models can achieve better accuracy with lower computation complexity than other methods.\\
\begin{table}
\caption{Results of ImageNet classification. We show the top-1 accuracy of each method under the same or closed FLOPs.}
\label{imagenet_results}
\begin{center}
\begin{tabular}[c]{c|ccc}
\hline
Network&Method&Acc@1&FLOPs\\
\hline
&Baseline$0.75\times$ &68.4\%&325M\\
&AMC\cite{b8} &70.5\%&285M\\
&SN\cite{b35} &69.5\%&325M\\
&MP\cite{b9} &70.4\%&281M\\
&AutoSlim\cite{b12} &69.1\%&325M\\
MobileNet V1&\textbf{SWP(ours)}&\textbf{70.9\%}&285M\\
\cline{2-4}
&Baseline$0.25\times$ &50.6\%&41M\\
&MP\cite{b9}&57.2\%&41M\\
&SN\cite{b35}&53.1\%&41M\\
&\textbf{SWP(ours)}&\textbf{60.3\%}&41M\\
\hline
\hline
&Baseline$0.75\times$&69.8\%&220M\\
&AMC \cite{b8}&70.8\%&220M\\
&MP\cite{b9} &71.2\%&220M\\
&SN\cite{b35}&68.9\%&209M\\
&AutoSlim\cite{b12}&72.4\%&207M\\
MobileNet V2&SWP(ours)&\textbf{73.4}\%&220M\\
\cline{2-4}
&Baseline$0.35\times$ &54.3\%&43M\\
&MP\cite{b9} &58.3\%&43M\\
&\textbf{SWP(ours)}&\textbf{60.7\%}&43M\\
\hline
\hline
&Baseline 1.0$\times$&76.6\%&4.1G\\
&Baseline 0.75$\times$ &74.8\%&2.3G\\
&SN\cite{b35} &74.9\%&2.3G\\
&MP\cite{b9}&75.4\%&2.3G\\
&AutoSlim\cite{b12}&75.6&2.3G\\
&AOFP-C1\cite{b43} &75.63\%&2.58G\\
&C-SGD-50\cite{b42} &74.54\%&1.7G\\
ResNet-50&ThiNet-50\cite{b41} &74.7\%&2.1G\\
&\textbf{SWP(ours)}&\textbf{76.1\%}&2.0G\\
\cline{2-4}
&Baseline 0.5$\times$&72.0\%&1.0\\
&SN\cite{b35}&72.5\%&1.0G\\
&ThiNet-30\cite{b41} &72.1\%&1.2G\\
&MP\cite{b9}&73.4\%&1.0G\\
&AutoSlim\cite{b12}&74.0\%&1.0G\\
&\textbf{SWP(ours)}&\textbf{75.6\%}&1.0G\\
\hline
\end{tabular}
\end{center}
\end{table}
\subsection{Pruning under Latency} 
More and more attention is paid to directly optimize the inference time on the target device. Without knowing the implementation details inside the device, SWP learns to prune channels according to the latency estimated from the device. To evaluate the realistic acceleration, we measure the forward time of the search for the best subnet on one 2080Ti GPU under the latency constraint. The results of MobileNet V1/V2 are shown in \textbf{\ref{inference-v1}}/\textbf{\ref{inference-v2}}. Under the same compression ratio and similar inference time, our method can obtain better top-1 accuracy.\\
\begin{table}
\caption{Comparison of the realistic inference time with MobileNet V1.} 
\label{inference-v1}
\begin{center}
\begin{tabular}[c]{c|c|c}
\hline
Ratio&Baseline/Pruned time(ms)&Baseline/Pruned Acc@1\\
\hline
$1\times$&0.65/-&70.9\%/-\\
$0.75\times$&0.51/0.51&68.4\%/70.2\%\\
$0.5\times$&0.35/0.35&63.3\%/67.5\%\\
$0.25\times$&0.21/0.20&49.8\%/59.9\%\\
\hline
\end{tabular}
\end{center}
\end{table}
\begin{table}
\caption{Comparison of the realistic inference time with MobileNet V2.} 
\label{inference-v2}
\begin{center}
\begin{tabular}[c]{c|c|c}
\hline
Ratio&Baseline/Pruned time(ms)&Baseline/Pruned Acc@1\\
\hline
$1\times$&0.72/-&71.7\%/-\\
$0.75\times$&0.53/0.52&69.8\%/71.2\%\\
$0.5\times$&0.41/0.41&65.4\%/69.3\%\\
$0.35\times$&0.33/0.32&60.3\%/64.8\%\\
\hline
\end{tabular}
\end{center}
\end{table}
\subsection{Combination with NAS}
We extend our effective method to NAS. Many NAS work\cite{b23,b24,b25} only search operations without considering the channel number. Here, we simultaneously take kernel size, layers and channels into account. Our network architectures are designed on MobileNet-V3\cite{b55} that consists of a stack with inverted bottleneck residual blocks(MB-Conv). The search space contains the channel number, kernel size and layer number of each stage. The detailed search space is shown in \textbf{\ref{mobilenetv3}}. \\
\begin{table}
\caption{MobileNetV3-based search space. \checkmark denotes the search space of the stage containing SE module\cite{b57}.} 
\label{mobilenetv3}
\begin{center}
\begin{tabular}[c]{c|c|c|c|c|c}
\hline
Stage&Operator&Ratio&Layers&Kernel Sizes&SE module\\
\hline
&Conv&1.0&1&3&\\
1&MBConv&0.1-1.0&1-2&3&\checkmark\\
2&MBConv&0.1-1.0&2-3&3&\checkmark\\
3&MBConv&0.1-1.0&2-3&3,5&\checkmark\\
4&MBConv&0.1-1.0&2-4&3,5&\checkmark\\
5&MBConv&0.1-1.0&2-6&3,5&\checkmark\\
6&MBConv&0.1-1.0&2-6&3,5&\checkmark\\
6&MBConv&0.1-1.0&1-2&3,5&\checkmark\\
&MBConv&1.0&1&1&\\
\hline
\end{tabular}
\end{center}
\end{table}
Several stage-wise supernets are trained on ImageNet dataset\cite{b15} using the same settings as MobileNet-V2. After the training, the optimal subnet is found based on distributed evolutionary algorithm. The retraining setting follows MobileNet-V3: RMSProp optimizer with decay 0.9 and momentum 0.9; batch size with 4096; initial learning rate 0.1 that decays by 0.01 every 3 epochs; dropout of 0.8 that only be applied in the fullnet. It is worth noting that a $3 \times 3$ kernel is central cropped from a $5 \times 5$ kernel and lower-index layers in each stage are always kept. Therefore, the weights also share on the kernel size and layer size dimension.\\
The performance of ours and previous SOTA methods with various search spaces are revealed in \textbf{\ref{NAS_result}}. Our models have 2.0\% and 1.7\% better top-1 accuracy than MobileNetV3-small and MobileNetV3-large with similar FLOPs, respectively. Compared with other mobile-setting NAS methods, our method outperforms them under wide-range FLOPs constraints because of more flexible search space. Specifically, our large-size model obtains a competitive result under fewer FLOPs compared with DNA\cite{b14} that also split a supernet into several stages but use a pretrained teacher network to supervise. It shows that the architecture of the teacher network has a huge impact on the effectiveness of distillation. We will discuss more details in \textbf{Sec 4.7}.
\begin{table}
\caption{Results of NAS methods on Imagenet.}
\label{NAS_result}
\begin{center}
\begin{tabular}[c]{c|cc}
\hline
Method&Acc@1&FLOPs\\
\hline
MobileNetV3-large\cite{b55}&75.2\%&219M\\
OFA\cite{b27}&76.0\%&230M\\
MNasNet\cite{b56} &75.2\%&315M\\
DNA-a\cite{b14} &77.1\%&348M\\
\textbf{SWPNas-large(ours)}&\textbf{76.9\%}&214M\\
\hline
\hline
MobileNetV3-small\cite{b55}&67.4\%&56M\\
Mnas-small\cite{b56} &64.9\%&65M\\
SWPNas-small&69.4\%&56M\\
\hline
\end{tabular}
\end{center}
\end{table}
\subsection{Visualization of Searched Models} 
Channel pruning models are visualized and some insights from the results are discussed. We compare our results with default channels and MetaPruning\cite{b9} on ResNet-50. In Figure \ref{6} (a-c), we show the channel number in the top, middle and bottom layers of bottleneck blocks on ResNet-50, respectively. Firstly, it is found that our method is prone to prune more channels from top layers compared with MetaPruning. It is noted that although top layers have a small number of channels, the output feature maps of the top layer can be extracted by the next middle layer where $kernel\ size = 3$. Hence, prune top layers can reduce computational complexity. Secondly, both our method and Metapruning keep more channels for downsampling layers because the feature map size is shrunk. Moreover, our method prunes fewer channels for bottom layers since the feature maps between the subnet and the fullnet should be as close as possible.\\
\subsection{Ablation Study} 
\label{sub:ablation_study}
\begin{figure}
\centering
\includegraphics[width=0.9\linewidth]{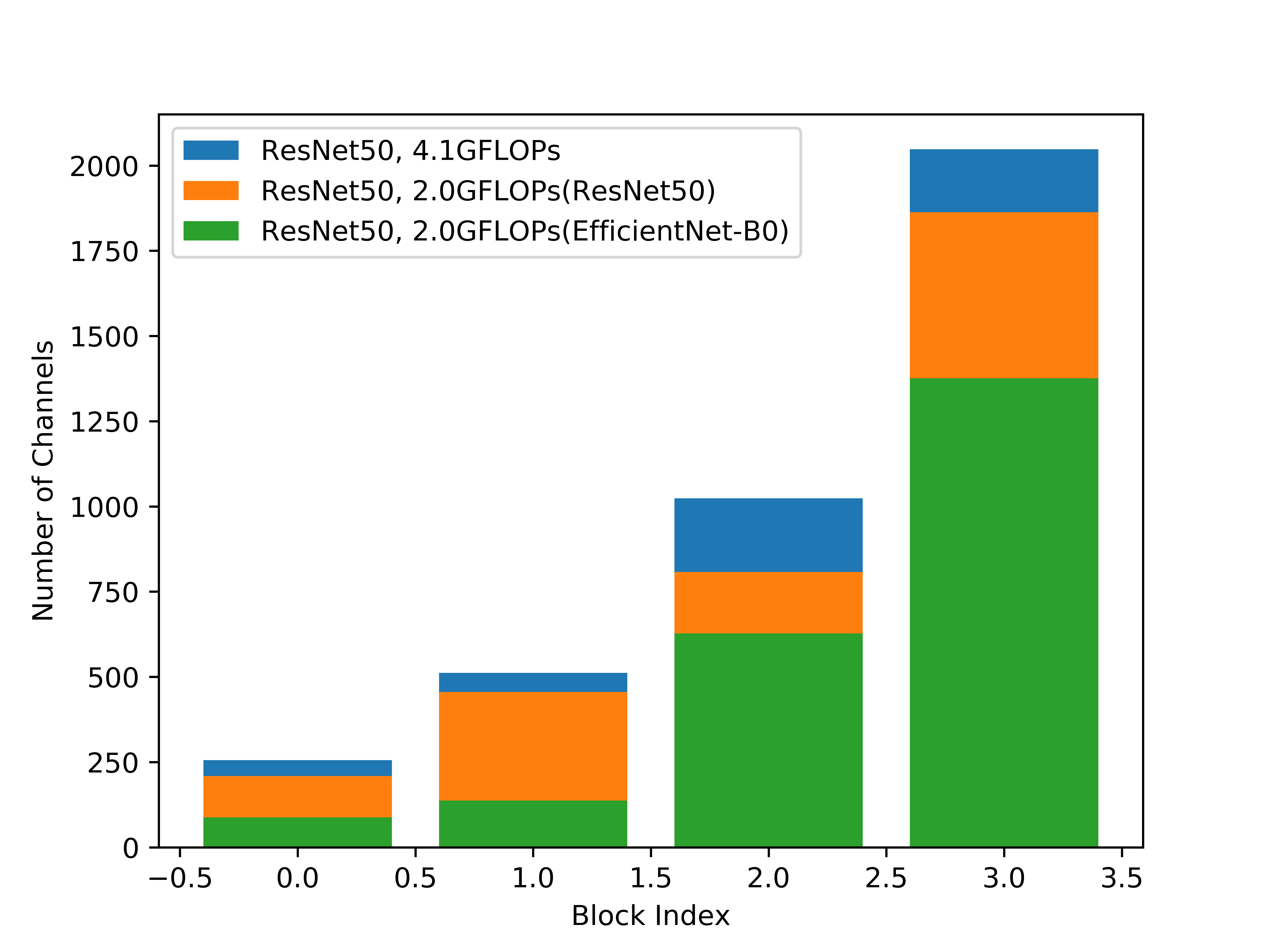}
\caption{The number of channels in bottom layers with different teachers.}
\label{7}
\end{figure}

\textbf{The choice of the teacher network.} The influence of the distillation strategy in ResNet-50 is analyzed. In our method, the teacher network and the student network are jointly trained by stage-wise inplace distillation. In Strategy 1, we use a pre-trained network that has the same architecture as fullnet to supervise the training of subnets. In Strategy 2, EfficientNet-B0\cite{b44} of which performance surpasses ResNet-50 with lower parameters are employed as the teacher network. The results are shown in Table \ref{distillation_strategies}. It is found that the performance of the model searched with inplace distillation method is almost the same as the one searched with a pre-trained method. Hence, the fullnet can supervise the training of the subnets well while training itself. It is unnecessary to spend a lot of time to obtain a pre-trained teacher model. Moreover, despite EfficientNet-B0 has outstanding performance compared with ResNet-50, the models searched with EfficientNet-B0 have worse performance. It may be caused by the large gap between the architectures of the teacher and the student networks.\\
To further figure out the reason why using ResNet-50 to search can achieve better performance of models than EfficientNet-B0, we visualize the channel number in bottom layers in each block. As shown in Figure \ref{7}, the model searched with EfficientNet-B0 keeps fewer channels. And EfficientNet-B0 has much fewer channels than ResNet-50. For example, EfficientNet-B0 has only 40 channels while ResNet-50 has 64 channels in the bottom layer of the first stage. Thus, the student does not need many channels to imitate the stage-wise feature maps generated by EfficientNet-B0. However, over-pruning the channels from the bottom layers will result in insufficient information transmission between the stages.\\
\begin{table}[t]
\caption{Comparison of Stage-wise pruning with different distillation strategies.}
\label{distillation_strategies}
\begin{center}
\begin{tabular}[c]{c|ccc}
\hline
Teacher&Student FLOPs&Acc@1\\
\hline
\hline
Ours&2.0G&76.1\%\\
&1.0G&75.6\%\\
\hline
\hline
Strategy 1&2.0G&76.1\%\\
&1.0G&75.6\%\\
\hline
\hline
Strategy 2&2.0G&75.6\%\\
&1.0G&73.1\%\\
\hline
\end{tabular}
\end{center}
\end{table}
\textbf{The stage number.} In previous experiments, the supernet is simply split into 4(ResNet) or 5(MobileNet series) stages according to downsampling. How many split stages are optimal to transfer knowledge from the teacher to the student? We set the stage number as 8, 4, 2, 1 for ResNet-50. It is seen from Table \ref{memory}, the performance of the model searched with 8 stages is almost the same as the one searched with 4 stages and surpasses the one searched with lower than 4 stages. Besides, since we need to save the intermediate feature maps of the teacher network to supervise the training of the student network, the more memory is required if larger stage number is set.\\
\begin{table}[t]
\caption{Comparison of Stage-wise pruning with different stage numbers. We evaluate the models with pruned ResNet-50 under 2.0G FLOPs constraint. The memory is tested under a 224$\times$224 image.}
\label{memory}
\begin{center}
\begin{tabular}[c]{c|ccc}
\hline
Number of stages&Acc@1&Memory\\
\hline
\hline
8&76.18\%&3M\\
4&76.12\%&1.5M\\
2&75.88\%&0.5M\\
1&74.53\%&0.1M\\
\hline
\end{tabular}
\end{center}
\end{table}
\subsection{Conclusion} 
\label{sub:conclusion}
In this work, we have presented stage-wise channel pruning. A stage-wise training process based on inplace distillation and a distributed evolutionary algorithm is proposed in this paper. It is found that the large search space causes low accuracy of the supernet. Hence, we split a supernet into several stage-wise supernets to degrade the complexity of the search space both in training and searching. The experiments show the effectiveness of our proposed method by delivering higher accuracy than the previous work on both CIFAR-10 and ImageNet dataset. The consistency of the proxy and actual subnet performance is greatly improved. Moreover, we extend our method to NAS and also achieve SOTA on the mobile setting. Experiments with various distillation strategies prove that inplace distillation can replace pre-trained distillation, thereby reducing the time to train a teacher network from scratch. We further discuss the impact of stage numbers on search results and found that splitting supernet according to downsampling is the best tradeoff between memory and accuracy. 

\section*{Acknowledgement}
\label{sub:acknowledgement}
We gratefully acknowledge the support of National Key R\&D Program of China (2018YFB1308400) and Natural Science Foundation of Zhejiang Province (NO. LY21F030018).

{\small
\bibliographystyle{ieee_fullname}
\bibliography{egbib}
}

\end{document}